\documentclass[conference]{IEEEtran}
\IEEEoverridecommandlockouts
\usepackage{cite}
\usepackage{amsmath,amssymb,amsfonts}
\usepackage{algorithmic}
\usepackage{graphicx}
\usepackage{textcomp}
\usepackage{xcolor}
\def\BibTeX{{\rm B\kern-.05em{\sc i\kern-.025em b}\kern-.08em
    T\kern-.1667em\lower.7ex\hbox{E}\kern-.125emX}}
\begin{document}

\title{MAGNet: Multi-agent Graph Network for Deep Multi-agent Reinforcement Learning
}

\author{\IEEEauthorblockN{ Aleksandra Malysheva}
\IEEEauthorblockA{JetBrains Research \\
National Research University\\
Higher School of Economics\\
St Petersburg, Russia \\
malyshevasasha777@gmail.com}
\and
\IEEEauthorblockN{Daniel Kudenko}
\IEEEauthorblockA{JetBrains Research \\
L3S Research Center\\
Leibniz University Hannover
\\
Hannover, Germany \\
daniel.kudenko@york.ac.uk}
\and
\IEEEauthorblockN{Aleksei Shpilman}
\IEEEauthorblockA{JetBrains Research \\
National Research University\\
Higher School of Economics\\
St Petersburg, Russia \\
aleksei@shpilman.com}}

\maketitle

\begin{abstract}
Over recent years, deep reinforcement learning has shown strong successes in complex single-agent tasks, and more recently this approach has also been applied to multi-agent domains. In this paper, we propose a novel approach, called MAGNet, to multi-agent reinforcement learning that utilizes a relevance graph representation of the environment obtained by a self-attention mechanism, and a message-generation technique. We applied our MAGnet approach to the synthetic predator-prey multi-agent environment and the Pommerman game and the results show that it significantly outperforms state-of-the-art MARL solutions, including Multi-agent Deep Q-Networks (MADQN), Multi-agent Deep Deterministic Policy Gradient (MADDPG), and QMIX. 
\end{abstract}


\section{Introduction}
A common difficulty of reinforcement learning in a multi-agent environment (MARL) is that in order to achieve successful coordination, agents require information about the relevance of environment objects to themselves and other agents. For example, in the game of Pommerman \cite{matiisen2018pommerman} it is important to know how relevant bombs placed in the environment are for teammates, e.g. whether or not the bombs can threaten them. While such information can be hand-crafted into the state representation for well-understood environments, in lesser-known environments it is preferable to derive it as part of the learning process. 

In this paper, we propose a novel method, named \mbox{MAGNet} (Multi-Agent Graph Network), to learn such relevance information in form of a relevance graph and incorporate this into the reinforcement learning process. The method works in two stages. In the first stage, a relevance graph is learned. In the second stage, this graph together with state information is fed to an actor-critic reinforcement learning network that is responsible for the decision making of the agent and incorporates message passing techniques along the relevance graph.

The contribution of this work is a novel technique to learn object and agent relevance information in a multi-agent environment, and incorporate this information in deep multi-agent reinforcement learning. 

We applied MAGNet to the synthetic predator-prey game, commonly used to evaluate multi-agent systems \cite{mordatch2018smag}, and the popular Pommerman \cite{matiisen2018pommerman} multi-agent environment. We achieved significantly better performance than state-of-the-art MARL techniques including MADQN \cite{egorov2016multi}, MADDPG \cite{lowe2017MADDPG}  and QMIX \cite{rashid2018qmix}. Additionally, we empirically demonstrate the effectiveness of utilized self-attention \cite{vaswani2017attenion}, graph sharing and message passing system.

\section{Deep Multi-Agent Reinforcement Learning}
In this section we describe the state-of-the-art deep reinforcement  learning  techniques  that  were  applied  to  multi-agent domains. The algorithms introduced below (MADQN, MADDPG, and QMIX) were also used as evaluation baselines in our experiments.

\subsection{Multi-agent Deep Q-Networks}



Deep Q-learning utilizes a neural network to predict Q-values of state-action pairs \cite{mnih2013playing}. This so-called deep Q-network is trained to minimize the following loss function:

\begin{equation} \label{eq:error-critic}
    L(\theta) =\mathbb{E}_{s, a \sim \rho(s)}[y - Q(s,a|\theta)]^2
\end{equation}

\begin{equation} \label{eq:policy}
    y = r + \gamma\max_{a'} Q^{past}(s',a')
\end{equation}

where $s'$ is the state we transition into by taking action $a$ in state $s$ and $r$ is the reward of that action, $\theta$ is the parameter vector of the current Q-function approximation. $a \sim \rho(s)$ denotes all actions that are permitted in state $s$. 

The Multi-agent Deep Q-Networks (MADQN,  \cite{egorov2016multi}) approach modifies this process for multi-agent systems by performing training in two repeated steps. First, they train agents one at a time, while keeping the policies of other agents fixed. When the agent is finished training, it distributes its policy to all of its allies as an additional environmental variable. 

\subsection{Multi-agent Deep Deterministic Policy Gradient}
When dealing with continuous action spaces, the MADQN method described above can not be applied. To overcome this limitation, the actor-critic approach to reinforcement learning was proposed \cite{sutton2000policy}. In this approach an actor algorithm tries to output the best action vector and a critic tries to predict the value function for this action.

Specifically, in the Deep Deterministic Policy Gradient (DDPG \cite{lillicrap2015DDPG}) algorithm two neural networks are used: $\mu(s)$ is the actor network that returns the action vector. 
$Q(s, a)$ is the critic network, that returns the $Q$ value, i.e. the value estimate of the action of $a$ in state $s$.

The gradient for the critic network can be calculated in the same way as the gradient for Deep Q-Networks described above (Equation \ref{eq:error-critic}). Knowing the critic gradient $\nabla_a Q$ we can then compute the gradient for the actor as follows:

\begin{equation} \label{eq:actor-gradient}
    \nabla_{\theta^{\mu}} J  = \mathbb{E}_{s}[\nabla_a Q(s,a| \theta^{q})|s = s_{t}, a = \mu(s|\theta^{\mu})]
\end{equation}

where $\theta^{q}$ and $\theta^{\mu}$ are parameters of critic and actor neural networks respectively, and $\rho^\pi(s)$ is the probability of reaching state $s$ with policy $\pi$.

The authors of \cite{lowe2017multi} proposed an extension of this method by creating multiple actors, each with its own critic, where each critic takes in the respective agent's observations and actions of all agents. This then constitutes the following value function for actor $i$:

\begin{equation}
\label{eq:maddpg}
\nabla_{\theta_i} J = E_{s, a \sim \pi^{\theta}}[\nabla_{\theta_{i}}log\pi_{i}(a_{i}|o_{i})Q^{\pi}_{i}(o_i, a_{1}, \dots, a_{N})]
\end{equation}

This Multi-agent Deep Deterministic Policy Gradient method showed the best results among widely used deep reinforcement learning techniques in continuous state and action space.

\subsection{QMIX}
Another recent promising approach to deep multi-agent reinforcement learning is the QMIX \cite{rashid2018qmix} method. It utilizes individual Q-functions for every agent and joint Q-function for a team of agents. The QMIX architecture consists of three types of neural networks.
\textbf{Agent networks} evaluate individual Q-functions for agents taking in the current observation and the previous action.
\textbf{Mixing network}  takes as input individual Q-functions from agent networks and a current state and then calculates a joint Q-function for all agents.
\textbf{Hyper networks} add an additional layer of complexity to the mixing network. Instead of passing the current state to the mixing network directly, hyper networks use it as input and calculate weight multipliers at each level of the mixing network. We refer the reader to the original paper for a more complete explanation \cite{rashid2018qmix}.

The authors empirically demonstrated on a number of RL domains that this approach outperforms both MADQN and MADDPG methods.

\section{MAGnet approach and architecture}
\label{sec:architecture}

\begin{figure*}[ht]
  \centering
    \includegraphics[width=0.7\linewidth]{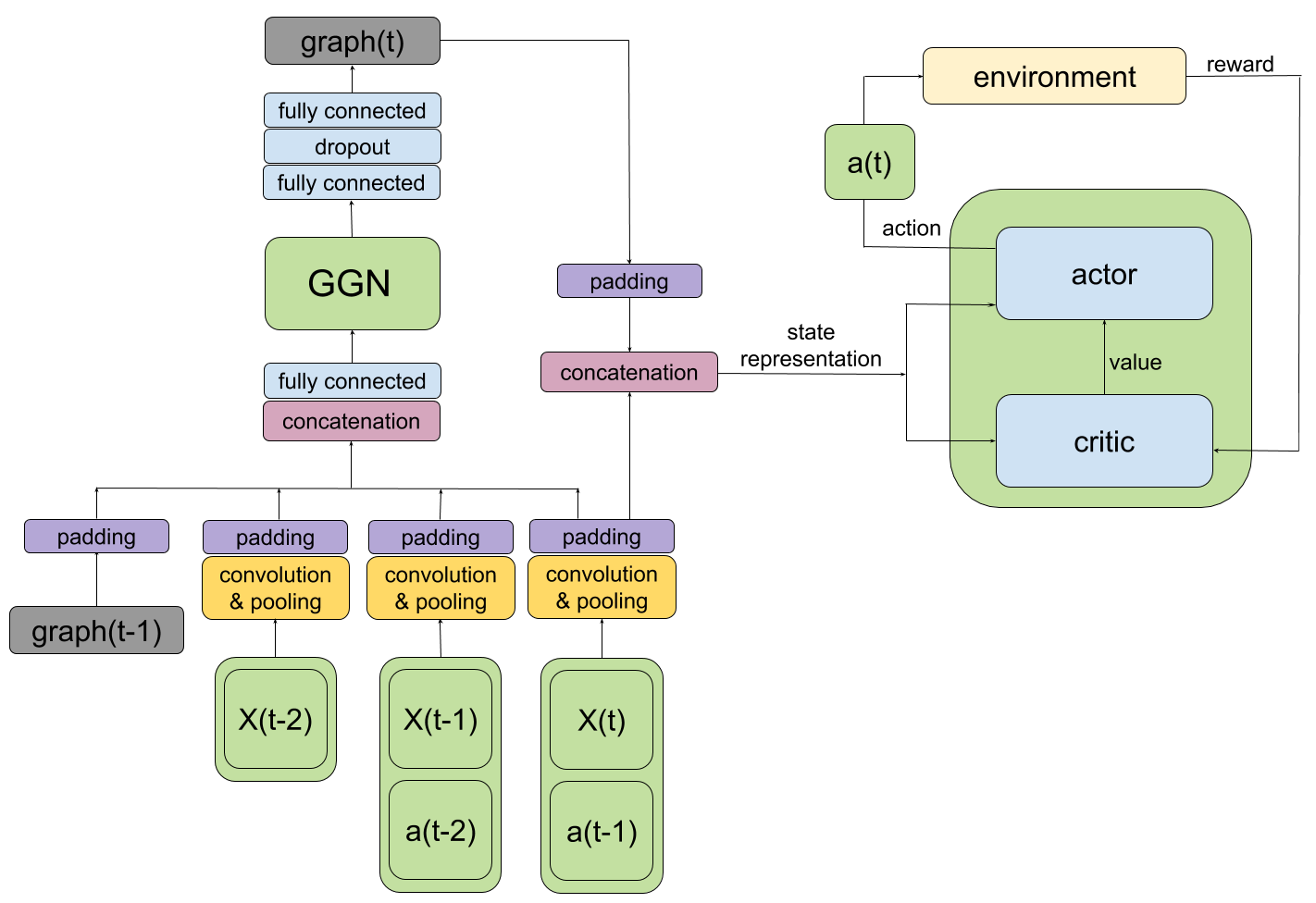}
    \caption{The overall network architecture of MAGNet. \textbf{Left} section shows the graph generation stage. \textbf{Right} part shows the decision making stage. $X(t)$ denotes the state of the environment at step $t$. $a(t)$ denotes the action taken by the agent at step $t$. GGN refers to Graph Generation Network.}
    \label{img:curr-network}
\end{figure*}

Figure \ref{img:curr-network} shows the overall network architecture of our MAGNet approach. The whole process can be divided into a relevance graph generation stage (shown in the left part) and a decision making stages (shown in the right part). In this architecture, the concatenation of the current state and the previous action forms the input of the model, and the output is the next action. The details of the two processes are described below.

\subsection{Relevance graph generation stage}\label{sec:graph-generation}
In the first part of our MAGNet approach, a neural network is trained to produce a relevance graph, which is represented as a numerical matrix $|A| \times (|A|+|O|)$, where $|A|$ is the number of agents and $|O|$ is a given maximum number of environment objects, e.g., bombs and walls in Pommerman. Weights for objects that are not present at the current time are set to 0. The relevance graph represents the relationship between agents and between agents and environment objects. The higher the absolute weight of an edge between an agent $a$ and another agent $b$ or object $o$ is, the more important $b$ or $o$ are for the achievement of agent $a$'s task. Every vertex $v$ of the graph has a type $b(v)$, defined by the user. Example types are "wall", "bomb", and "agent". Types are user-defined and are used in the message generation step (see below). The graph is generated by MAGNet from the current and previous state together with the respective actions.  

\section{Relevance graph visualization} \label{sec:app:graph}

To generate this relevance graph, we train a neural network via back-propagation to output a graph representation matrix. The input to the network are the current and the two previous states (denoted by $X(t)$, $X(t-1)$, and $X(t-2)$ in Figure~\ref{img:curr-network}), the two previous actions (denoted by $a(t-1)$ and $a(t-2)$), and the relevance graph produced at the previous time step (denoted by $graph(t-1)$). For the first learning step (i.e. $t=0$), the input consists out of three copies of the initial state, no actions, and a random relevance graph. The inputs are passed into a convolution and pooling layer, followed by a padding layer, and then concatenated and passed into fully connected layer and finally into the graph generation network (GGN). In this work we implement GGN as either a multilayer perceptron (MLP) or a self-attention network, which uses an attention mechanism to catch long and short term time-dependencies. We present the results of both implementations in Table~\ref{tbl:modules}. The self-attention network is an analogue to a recurrent network, such as an LSTM, but takes much less time to compute \cite{vaswani2017attenion}. The result of the GGN is fed into a two-layer fully connected network with dropout, which produces the relevance graph matrix.

The loss function for the back-propagation training is defines as follows:
\begin{equation}\label{eq:graph}
    L = ||W^{t} - W^{t-1}||_2^2
\end{equation}
 
The loss function is based on the squared difference between weights of edges in the current graph  $W^{t}$ and the one generated in the previous state $W^{t-1}$. We could train the graph without this loss function, and instead with just the loss function of the decision making stage backpropagated to the graph generation stage. However, we found that this lowers the performance (see Figure~\ref{img:best}).
 Both Pommerman and predator-prey environments have these default agents. However, we found out that the better way to train MAGNet is to first pre-train the graph generation and then add the agent networks (see also Section~\ref{sec:pt}). There are two alternatives for training relevance graphs: (1) train individual relevance graphs for every agent, or (2) train one shared graph (GS) that is the same for all agents on the team. We performed experiments to determine which way is better (see Table~\ref{tbl:modules}).

\subsection{Decision making stage}\label{sec:message}

The agent AI responsible for decision making is represented as a neural network whose inputs are accumulated messages and the current state of the environment. The output of the network is an action to be executed. 
This action is computed in 4 steps through a message passing system.  


In the first step, individual (i.e. location-specific) observations of agents and objects are pre-processed by a neural network into an information vector (represented as a numerical vector). This neural network is initialized randomly and trained during the overall learning process using the same global loss function. 

In the second step, a neural network (also trained) is taking the information vector of an agent, and maps it into a message (also a numerical vector), one for each connected vertex type in the relevance graph. This message is multiplied by the weight of the corresponding edge and passed to the respective vertices. 

In the third step, each agent or object in the relevance graph updates its information vector, also using a trained network, based on the incoming messages and the previous information vector. Steps 2 and 3 are repeated a given number of times, in our experiments five times. Finally, in the last step, the final message received by the agent, together with the current state information is mapped by a trained decision making network into an action. 

Since the message passing system outputs an action, we view it as an actor in the DDPG actor-critic approach \cite{lillicrap2015DDPG}, and train it accordingly. A more formal description of this decision making stage is as follows.

\begin{enumerate}
    \item \textbf{Initialization of information vector.}
    Each vertex $v$ has an initialization network $\textit{MLP}^{b(v)}_{init}$ associated with it according to its type $b(v)$  that takes as input the current individual observation $O_v$ and outputs initial information vector $\mu^0_v$ for each vertex. 
    \begin{equation}
    \label{eq:init_info_v}
        \mu^0_v = \textit{MLP}^{b(v)}_{init} (O_v)
    \end{equation}
    
    \item \textbf{Message generation.}
    Message generation performs in iterative steps. At message generation step $\tau+1$ (not to be confused with environmental time $t$) the message networks $\textit{MLP}^{c{(v,u)}}_{msg}$ compute output messages for every edge $(v,u) \in E$ based on the type of the edge $c(v,u)$, which are then multiplied by the weight of the corresponding edge in the relevant graph. 
    \begin{equation}
    \label{eq:out_v}
        \textit{msg}^{\tau}_{(v, u)} = w_{(v,u)}\textit{MLP}^{c{(v,u)}}_{msg} (\mu^{\tau}_v)
    \end{equation}
    
    \item \textbf{Message processing.} 
    The information vector $\mu^{\tau+1}_v$ at the message propagation step $\tau$ is updated by an associated update network $\textit{LSTM}^{b(v)}_{up}$, according to its type $b(v)$. The network takes as input a sum of all incoming message vectors and the information vector $\mu^{\tau}_v$ at the previous step.
    \begin{equation}
    \label{eq:processing}
        \mu^{\tau+1}_v = \textit{LSTM}^{b(v)}_{up}(\mu^{\tau}_v, \sum{msg^\tau_{(*,v)}})
    \end{equation}
    
    \item \textbf{Choice of action.}
    All vertices that are associated with agents have a decision network $\textit{MLP}^{b(v)}_{\textit{choice}}$ which takes as an input its final information vector $\mu^T_v$ and computes the mean of the action of the Gaussian policy. 
    
    \begin{equation}
    \label{eq:choise_nn}
        a_v = \textit{MLP}_{\textit{choice}}(\mu^{T}_v)
    \end{equation}

\end{enumerate}

\section{Experiments}

\subsection{Environments}

In this paper, we use two popular multi-agent benchmark environments for testing, the synthetic multi-agent predator-prey game \cite{mordatch2018smag}, and the Pommerman game \cite{matiisen2018pommerman}. 

In the predator-prey environment, the aim of the predators is to kill faster moving prey in 500 iterations. The predator agents must learn to cooperate in order to surround and kill the prey. Every prey has a health of 10. A predator moving within a given range of the prey lowers the prey's health by 1 point per time step. Lowering the prey health to 0 kills the prey. If even one prey survives after 500 iterations, the prey team wins. Random obstacles are placed in the environment at the start of the game.

The Pommerman game is a popular environment which can be played by up to 4 players. The multi-agent variant has 2 teams of 2 players each. This game has been used in recent competitions for multi-agent algorithms, and therefore is especially suitable for a comparison to state-of-the-art techniques. 

In Pommerman, the environment is a grid-world where each agent can move in one of four directions, place a bomb, or do nothing. A grid square is either empty (which means that an agent can enter it), wooden, or rigid. Wooden grid squares can not be entered, but can be destroyed by a bomb (i.e. turned into clear squares). Rigid squares are indestructible and impassable. When a wooden square is destroyed, there is a probability of items appearing, e.g., an extra bomb, a bomb range increase, or a kick ability. Once a bomb has been placed in a grid square it explodes after 10 time steps. The explosion destroys any wooden square within range 1 and kills any agent within range 4. If both agents of one team die, the team loses the game and the opposing team wins. The map of the environment is randomly generated for every episode. 

The game has two different modes: free for all and team match. Our experiments were carried out in the team match mode in order to evaluate the ability of MAGnet to exploit the discovered relationships between agents (e.g. being on the same team).

We represent states in both environments as a $D\times D\times M$ tensor $S$, where $D\times D$ are the dimensions of the field and $M$ is the maximum possible number of objects. $S[i,j,k]=1$ if object $k$ is present in the $[i,j]$ space and is 0 otherwise. The predator-prey state is a $64\times 64\times 20$ tensor, and the Pommerman state is $11\times 11\times 30$.

\begin{figure}[ht]
\centering     
\includegraphics[width=0.9\linewidth]{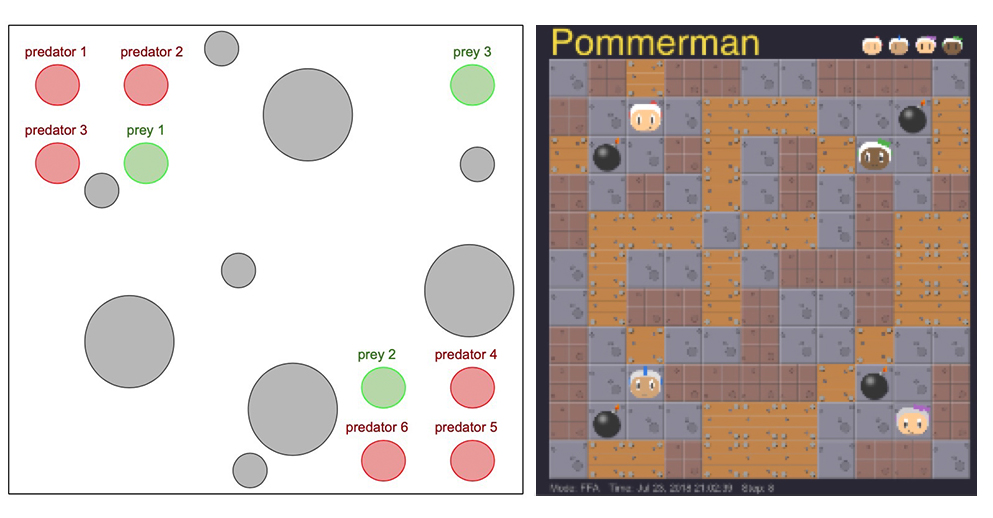}
\caption{The synthetic predator-prey (left) and the Pommerman game environment (right).}
\label{img:predator-prey}
\end{figure}

Figure~\ref{img:predator-prey} shows both test environments.

\subsection{Evaluation Baselines}

\begin{figure*}[ht]
\centering     
\includegraphics[width=0.9\linewidth]{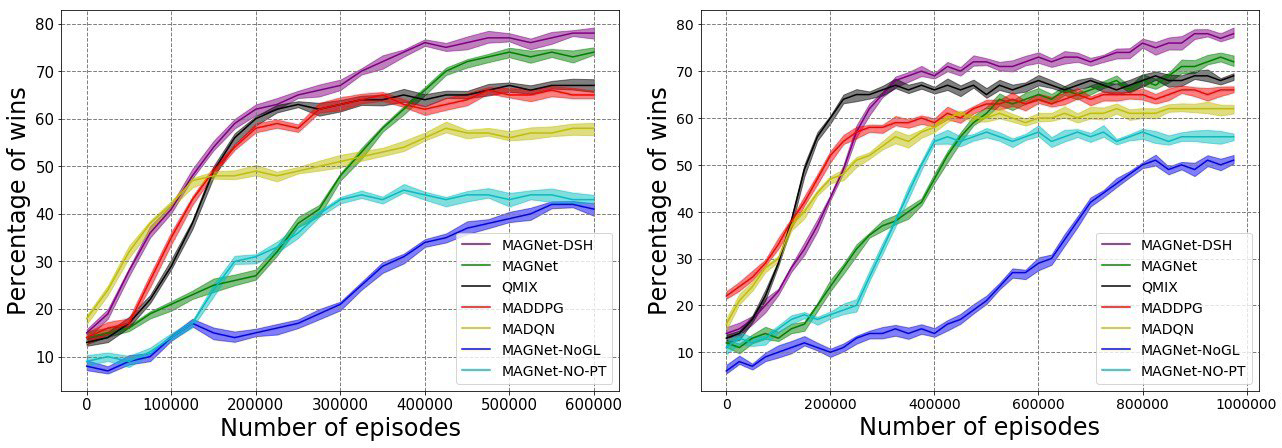}
\caption{MAGNet variants compared to state-of-the-art MARL techniques in the predator-prey (left) and Pommerman (right) environments. MAGNet-DSH refers to MAGNet with domain specific heuristics (Section~\ref{sec:dsh}). MAGNet-NoGL refers to MAGNet trained without the graph generation loss function (Equation~\ref{eq:graph}), but only with the final loss function of the decision making stage. MAGNet-NO-PT refers to MAGNet with no pre-training for the graph generating network (Section~\ref{sec:pt} ). Every algorithm trained by playing against a default environment agent for a number of games (episodes) and a respective win percentage is shown. Default agents are provided by the environments. Shaded areas show the 95\% confidence interval from 20 runs.}
\label{img:best}
\end{figure*}



In our experiments, we compare the proposed method with state-of-the-art reinforcement learning algorithms in the two environments described above. 
Figure~\ref{img:best} shows a comparison with the MADQN \cite{egorov2016multi}, MADDPG \cite{lowe2017MADDPG} and  QMIX \cite{rashid2018qmix} algorithms. Each of these algorithms were trained by playing a number of games (i.e. episodes) against the default AI provided with the games, and the respective win rates are shown. All graphs display a 95\% confidence interval over 20 runs to show statistical significance. 

The parameters for the \textbf{MADQN} the baselines were set as follows through parameter exploration. The network for the predator-prey environment consists of seven convolutional layers with 64 5x5 filters in each layer followed by five fully connected layers with 512 neurons each with residual connections \cite{he2016resnet} and batch normalization \cite{ioffe2015batch} that takes an input an 128x128x6 environment state tensor and one-hot encoded action vector (a padded 1x5 vector) and outputs a Q-function for that state-action pair. Since the output of a DQN is discrete, but the predator-prey environment requires a continuous action, the agents employ only two speeds and 10 directions. The network for Pommerman consists of five convolutional layers with 64 3x3 filters in each layer followed by three fully connected layers with 128 neurons each with residual connections and batch normalization that takes an input an 11x11x4 environment state tensor and one-hot encoded action vector (a padded 1x6 vector) that are provided by the Pommerman environment and outputs a Q-function for that state-action pair.

For our implementation of \textbf{MADDPG} we used a multilayer perceptron (MLP) with 3 fully connected layers with 512-128-64 neurons for both the actor and critic for the predator-prey game, and 5 fully connected layer with 128 neurons in each layer for the critic and a 3 layer network with 128 neurons in each layer for the actor in Pommerman. 

Parameter exploration for \textbf{QMIX} led to the following settings for both environments. All agent networks are DQNs with a recurrent layer of a Gated Recurrent Unit (GRU \cite{chung2015gated}) with a 64-dimensional hidden state. 
The mixing network consists of a single hidden layer of 32 neurons. Since the output of MADDPG and QMIX is continuous, but Pommerman expects a discrete action, we discretized it. 

As in the original QMIX paper \cite{rashid2018qmix}, we decrease learning rate linearly from 1.0 to 0.05 over the first 50k time steps and than keep it constant. As can be seen from Figure~\ref{img:best}, our MAGnet approach significantly outperforms current state-of-the-art algorithms.

\subsection{MagNet network training}\label{sec:magnet-training}

In both environments we first trained the graph generating network on 50,000 episodes with the same parameters and with the default AI as the decision making agents. Both predator-prey and Pommerman environments provide these default agents. After this initial training, the default AI was replaced with the learning decision making AI described in section~\ref{sec:architecture}. All learning graphs show the training episodes starting after this replacement.

Table ~\ref{tbl:modules} shows results for different MAGNet variants in terms of achieved win percentage against a default agent after 600,000 episodes in the predator-prey game and a 1,000,000 episodes in the game of Pommerman. The MAGNet variants are differing in the complexity of the approach, starting from the simplest version which takes the learned relevance graph as a direct addition to the input, to the version incorporating message generation, graph sharing, and self-attention. The table clearly shows the benefit of each extension.

\begin{table}[ht]
\caption{Influence of different modules on the performance of the MAGnet model.}
\label{tbl:modules}
\begin{center}
\begin{tabular}{| p{1.3cm} | p{1.2cm} | p{1.4cm} | p{1.4cm} | p{1.4cm} |}
\hline
\multicolumn{3}{| c |}{\textbf{MAGnet modules}}  & \textbf{Win \% PP} & \textbf{Win \% PM}\\ 
\cline{1-3}
\textbf{SA} & \textbf{GS} & \textbf{MG} & & \\
\hline
+ & + & + & $74.2\pm 1.2$ & $76.3\pm 0.7$\\ \hline
+ & + & - & $61.3\pm 0.9$ & $56.7\pm 1.8$\\ \hline
+ & - & + & $63.2\pm 1.3$ & $62.4\pm 1.7$\\ \hline
+ & - & - & $43.3\pm 1.5$ & $54.5\pm 2.6$\\ \hline
- & + & + & $69.3\pm 1.5$ & $67.1\pm 1.9$\\ \hline
- & + & - & $39.3\pm 2.0$ & $52.0\pm 1.7$\\ \hline
- & - & + & $41.5\pm 1.4$ & $45.2\pm 3.6$\\ \hline
- & - & - & $25.1\pm 2.3$ & $32.7\pm 5.9$\\ \hline
\end{tabular}
\end{center}
\end{table}


Each of the three extensions with their hyper-parameters are described below: 

\begin{itemize}
\item Self-attention (\textbf{SA}). We can train the Graph Generating Network (GGN) as a simple multi-layer perceptron (number of layers and neurons was varied, and a network with 3 fully connected layers 512-128-128 neurons achieved the best result). Alternatively, we can train it as a self-attention encoder part of the Transformer network (\textbf{SA}) layer \cite{vaswani2017attenion} with default parameters.

\item Graph Sharing (\textbf{GS}): relevance graphs were trained individually for agents, or in form of a shared graph for all agents on one team.

\item Message Generation (\textbf{MG}): the message generation module was implemented as either an MLP or a message generation (MG) architecture, as described in Section \ref{sec:message}.
\end{itemize}






\subsection{MAGNet parameters}
We define vertex types $b(v)$ and edge types $c(e)$ in relevance graph as follows:

$b(v) \in \{0, 1, 2, 3\}$ in case of predator-prey game that corresponds to: "predator on team 1 (1, 2, 3)", "predator on team 2 (4, 5, 6)", "prey", "wall". Every edge has a type as well: $c(e) \in \{0, 1, 2\}$, that corresponds to “edge between predators within one team”, “edge between predators from different teams” and “edge between the predator and the object in the environment or prey”.

$b(v) \in \{0, 1, 2, 3, 4, 5, 6\}$ in case of Pommerman game that corresponds to: "ally", "enemy", "placed bomb" (about to explode), "increase kick ability", "increase blast power", "extra bomb" (can be picked up). Every edge has a type as well: $c(e) \in \{0, 1\}$, that corresponds to “edge between the agents” and “edge between the agent and the object in the environment”.

We tested the MLP and message generation network with a range of hyper-parameters, choosing the best one. In the predator-prey game, the MLP has 3 fully connected layers with 512-512-128 neurons, while the message generation network has 5 layers with 512-512-128-128-32 neurons. For the Pommerman environment, the MLP has 3 fully connected layers 1024-256-64 neurons, while the message generation network has 2 layers with 128-32 neurons. In both domains 5 message passing iterations showed the best result.

Dropout layers were individually optimized by grid search in the [0, 0.2, 0.4] space. We tested two convolution sizes: [3x3] and [5x5]. [5x5] convolutions showed the best result. A Rectified Linear Unit (ReLU) transformation was used for all connections.



\subsection{No pre-training}
\label{sec:pt}

With regards to pre-training of the graph generating network we need to answer the following questions. First, we need to determine whether or not it is feasible to train the network without an external agent for pre-training. In other words, can we simultaneously train both the graph generating network and the decision making networks from the start. Second, we need to demonstrate whether pre-training of a graph network improves the result. 

To answer this question, we performed experiments without the pre-training of the graph network. Figure~\ref{img:best} shows the results of those experiments (MAGNet-NO-PT). As can be seen, the network indeed can learn without pre-training, but pre-training significantly improves the results. This may be due to decision making error influencing the graph generator network in a negative way.

\subsection{Domain specific heuristics}
\label{sec:dsh}

We also performed experiments to see whether or not additional knowledge about the environment can improve the results of our method. To incorporate this knowledge, we change the loss function for graph generation in the following manner.

\begin{equation}
    \label{eq:loss_graph}
    L = ||W^{t} - W^{t-1}||_2^2 + \sum_{\xi_{(v,u)} \in \Xi^t}(w^{t}_{(v,u)} - s(\xi_{(v,u)}))^2
\end{equation}

The first component is the same: it is based on the squared difference between weights of edges in the current graph  $W^{t}$ and the one generated in the previous state $W^{t-1}$. The second iterates through events $\Xi^t$ at time $t$ and calculates the square difference between the weight of edge $(v,u)$ that is involved in event $\xi_{(v,u)}$ and the event weight $s(\xi_{(v,u)})$.

For example, in the Pommerman environment we set $s(\xi)$ corresponding to our team agent killing an agent from the opposite team to 100, and the $s(\xi)$ corresponding to an agent picking up a bomb to 25. In the predator-prey environment, if a predator kills a prey, we set the event's weight to 100. If a predator only wounds the prey, the weight for that event is set to 50.

As can be seen in Figure~\ref{img:best} (line MAGNet-DSH), the model that uses this domain knowledge about the environment trains faster and performs better. It is however important to note that the MAGNet network without any heuristics still outperforms current state-of-the-art methods. For future research we consider creating a method for automatic assignment of the event weights.  


\section{Conclusion}
In this paper we presented a novel method, MAGNet, for deep multi-agent reinforcement learning incorporating information on the relevance of other agents and environment objects to the RL agent. We also extended this basic approach with various optimizations, namely self-attention, shared relevance graphs, and message generation. The MAGNet variants were evaluated on the popular predator-prey and Pommerman game environments, and compared to state-of-the-art MARL techniques. Our results show that MAGNet significantly outperforms all competitors.

\bibliographystyle{unsrt}
\bibliography{references}

\begin{thebibliography}{10}

\bibitem{matiisen2018pommerman}
Tambet Matiisen.
\newblock Pommerman baselines, 2018.

\bibitem{mordatch2018smag}
Igor Mordatch and Pieter Abbeel.
\newblock Emergence of grounded compositional language in multi-agent
  populations.
\newblock In {\em AAAI Conference on Artificial Intelligence}, 2018.

\bibitem{egorov2016multi}
Maxim Egorov.
\newblock Multi-agent deep reinforcement learning, 2016.

\bibitem{lowe2017MADDPG}
Ryan Lowe, Yi~Wu, Aviv Tamar, Jean Harb, OpenAI~Pieter Abbeel, and Igor
  Mordatch.
\newblock Multi-agent actor-critic for mixed cooperative-competitive
  environments.
\newblock In {\em Advances in Neural Information Processing Systems}, pages
  6379--6390, 2017.

\bibitem{rashid2018qmix}
Tabish Rashid, Mikayel Samvelyan, Christian~Schroeder de~Witt, Gregory
  Farquhar, Jakob Foerster, and Shimon Whiteson.
\newblock Qmix: monotonic value function factorisation for deep multi-agent
  reinforcement learning.
\newblock {\em arXiv preprint arXiv:1803.11485}, 2018.

\bibitem{vaswani2017attenion}
Ashish Vaswani, Noam Shazeer, Niki Parmar, Jakob Uszkoreit, Llion Jones,
  Aidan~N Gomez, {\L}ukasz Kaiser, and Illia Polosukhin.
\newblock Attention is all you need.
\newblock In {\em Advances in Neural Information Processing Systems}, pages
  5998--6008, 2017.

\bibitem{mnih2013playing}
Volodymyr Mnih, Koray Kavukcuoglu, David Silver, Alex Graves, Ioannis
  Antonoglou, Daan Wierstra, and Martin Riedmiller.
\newblock Playing atari with deep reinforcement learning.
\newblock {\em arXiv preprint arXiv:1312.5602}, 2013.

\bibitem{sutton2000policy}
Richard~S Sutton, David~A McAllester, Satinder~P Singh, and Yishay Mansour.
\newblock Policy gradient methods for reinforcement learning with function
  approximation.
\newblock In {\em Advances in neural information processing systems}, pages
  1057--1063, 2000.

\bibitem{lillicrap2015DDPG}
Timothy~P Lillicrap, Jonathan~J Hunt, Alexander Pritzel, Nicolas Heess, Tom
  Erez, Yuval Tassa, David Silver, and Daan Wierstra.
\newblock Continuous control with deep reinforcement learning.
\newblock {\em arXiv preprint arXiv:1509.02971}, 2015.

\bibitem{lowe2017multi}
Ryan Lowe, Yi~Wu, Aviv Tamar, Jean Harb, OpenAI~Pieter Abbeel, and Igor
  Mordatch.
\newblock Multi-agent actor-critic for mixed cooperative-competitive
  environments.
\newblock In {\em Advances in Neural Information Processing Systems}, pages
  6379--6390, 2017.

\bibitem{he2016resnet}
Kaiming He, Xiangyu Zhang, Shaoqing Ren, and Jian Sun.
\newblock Deep residual learning for image recognition.
\newblock In {\em Proceedings of the IEEE conference on computer vision and
  pattern recognition}, pages 770--778, 2016.

\bibitem{ioffe2015batch}
Sergey Ioffe and Christian Szegedy.
\newblock Batch normalization: Accelerating deep network training by reducing
  internal covariate shift.
\newblock {\em arXiv preprint arXiv:1502.03167}, 2015.

\bibitem{chung2015gated}
Junyoung Chung, Caglar Gulcehre, Kyunghyun Cho, and Yoshua Bengio.
\newblock Gated feedback recurrent neural networks.
\newblock In {\em International Conference on Machine Learning}, pages
  2067--2075, 2015.

\end{thebibliography}

\end{document}